\documentclass[10pt,twocolumn,letterpaper]{article}

\usepackage[pagenumbers]{cvpr} %

\usepackage{multirow}
\usepackage{tabularx}

\usepackage{microtype}

\definecolor{cvprblue}{rgb}{0.21,0.49,0.74}
\usepackage[pagebackref,breaklinks,colorlinks,allcolors=cvprblue]{hyperref}

\def\confName{CVPR}
\def\confYear{2027}

\title{LoopVGGT: Training-Free Looped Vision Geometry Transformers}

\author{
Yifei Liu$^{1*}$ \quad
Jiahui Sun$^{1*}$ \quad
Huaxi Huang$^{1}$ \quad
Hongjie Zhang$^{1}$ \quad
Xiao Sun$^{1}$ \quad
Zhihang Zhong$^{2\dagger}$\\[4pt]
$^{1}$Shanghai Artificial Intelligence Laboratory \qquad
$^{2}$Shanghai Jiao Tong University\\[2pt]
{\small $^{*}$Equal contribution \qquad $^{\dagger}$Corresponding author}
}

\begin{document}
\maketitle
\begin{abstract}
Can the geometric predictions of pretrained feed-forward models be improved without retraining or fine-tuning?
We investigate training-free looped inference across five geometric foundation models: VGGT, Pi3, Depth Anything~3, MapAnything, and VGGT-$\Omega$.
We introduce LoopVGGT, an architecture-aware framework that replays a contiguous block window with damped updates while preserving each backbone's native computation graph, including paired within-view and cross-view attention, specialized or scale tokens, and dense head interfaces.
For VGGT, we further incorporate a dual-stream cache update so looping extends to windows containing intermediate DPT readout layers.
Across nine static and dynamic benchmarks with one fixed configuration per model, looped inference yields broad aggregate improvements and substantial dataset-level gains, raising VGGT's HiRoom reconstruction F1 from 53.79 to 60.43 and improving all reported reconstruction metrics for MapAnything.
We further characterize task-dependent trade-offs, task-specialist alternatives, and retrospective single-source transfer across multiple backbones.
Diagnostic analyses on HiRoom show that higher intra-window feature similarity correlates with larger reconstruction gains in VGGT and Pi3, downstream feature-direction deviation separates beneficial from harmful VGGT windows, and consistent cache updates enable effective looping even across internal DPT readout layers.
Our code is publicly available at \url{https://github.com/Visionary-Laboratory/LoopVGGT}.
\end{abstract}

\section{Introduction}
\label{sec:intro}
Multi-view geometry has increasingly become a feed-forward prediction problem.
Given a set of images, pretrained transformers can jointly recover camera poses, depth, and 3D structure without optimizing a scene-specific representation
\citep{wang2025vggt,wang2025pi3,lin2025depthanything3,keetha2025mapanything,wang2026vggtomega}.
However, training these reconstruction foundation models requires large-scale datasets and substantial computational resources \citep{wang2025vggt,lin2025depthanything3}.
This investment motivates the central question of this paper: \emph{can we improve the geometric predictions of existing feed-forward models without retraining or fine-tuning them?}

\begin{figure}[t]
\centering
\includegraphics[width=0.92\linewidth]{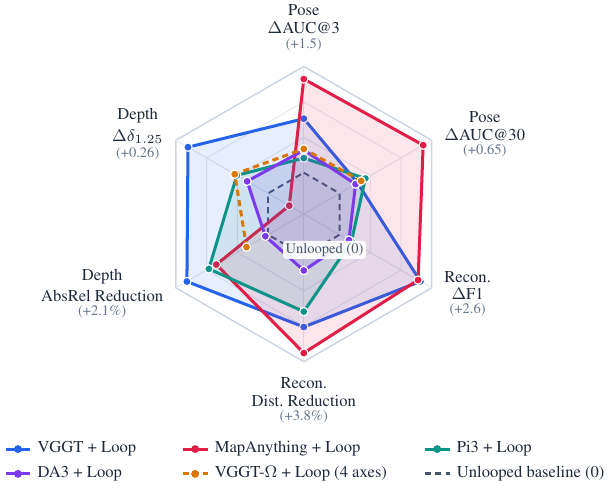}
\caption{Cross-benchmark aggregate gains of training-free looping over each model's unlooped baseline (dashed inner hexagon) under one fixed shared configuration per model (Table~\ref{tab:aggregate}). Each axis uses a linear scale anchored at 0 with its outer-ring bound in parentheses; outward indicates improvement. VGGT-$\Omega$ evaluates camera pose and depth (four axes).}
\label{fig:radar_overview}
\end{figure}

\begin{figure*}[t]
\centering
\includegraphics[width=0.92\textwidth]{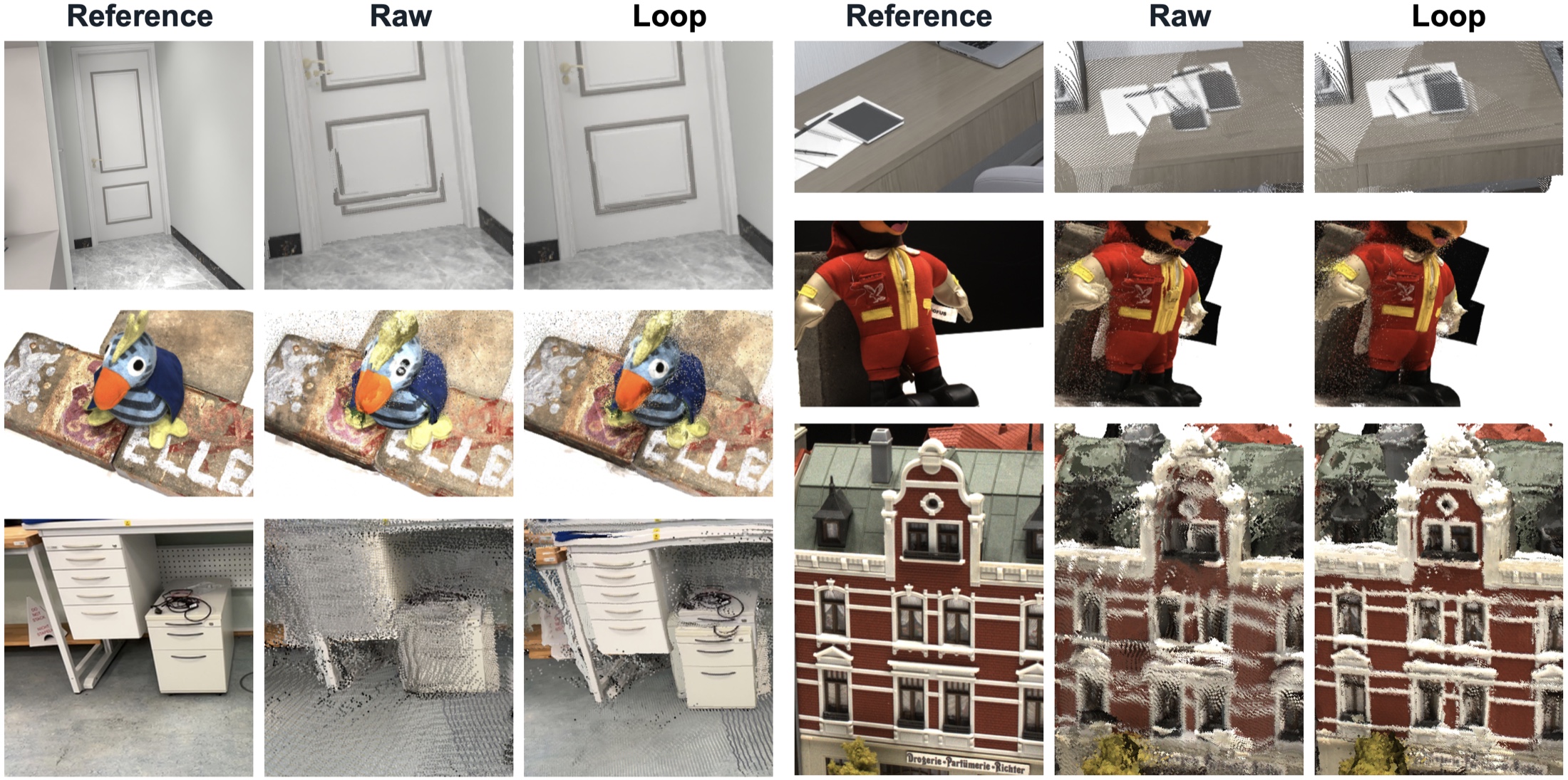}
\caption{Qualitative reconstruction examples. From top to bottom in the left group, then top to bottom in the right group, the models are VGGT, Pi3, MapAnything, VGGT, DA3, and VGGT-$\Omega$. Triplets show input RGB, raw, and loop, with identical observation cameras and rendering settings for each raw/loop pair. Looping improves structural coherence across diverse scenes; input photographs serve as visual references rather than pixel-aligned ground truth.}
\label{fig:qualitative}
\end{figure*}

Looped transformers reuse the same layers across depth \citep{dehghani2019universal,giannou2023looped,yang2024looped}, and training with this recurrent structure has shown benefits for algorithmic generalization and language-model reasoning \citep{fan2025looped,saunshi2025reasoning,geiping2025scaling,zhu2025ouro}.
Recent work further shows that looping can improve frozen language models without additional training \citep{chen2026trainingfree,lys2026innerloop}.
Motivated by these findings, we investigate whether repeating a contiguous group of layers (a \emph{loop window}) within pretrained reconstruction models can also improve multi-view geometric predictions.

Extending training-free looping from single-stream language models to multi-view geometry introduces two structural considerations.
First, geometry backbones rely on heterogeneous computation graphs that must remain intact during replay, including alternating within-view and cross-view attention pairs, special communication tokens, staged single- and multi-view blocks, and scene-level scale tokens.
Second, dense prediction heads such as DPT \citep{ranftl2021dpt} read intermediate features cached across multiple backbone depths rather than only the final layer, requiring either cache-consistent window placement or consistent updates to the cached states when a window spans those readout layers.
Moreover, the same loop can affect camera pose, depth, and surface reconstruction differently, making its benefits task-dependent.

We develop \textbf{LoopVGGT}, an architecture-aware framework for applying training-free looping across five pretrained geometry backbones.
Our wrapper repeats native attention blocks or pairs with damped updates while preserving special tokens, stage structure, and dense-head interfaces.
To extend looping to VGGT windows that contain intermediate readout layers, we additionally introduce a dual-stream cache update (Equation~\ref{eq:cache}) that evolves both the frame and global features saved for dense prediction.
We evaluate the framework across VGGT, Pi3, Depth Anything~3 (DA3), MapAnything, and VGGT-$\Omega$ on nine benchmarks spanning real and synthetic data, indoor and outdoor scenes, and static and dynamic environments.
Each model uses one fixed loop configuration across its supported benchmarks, complemented by retrospective single-source held-out transfer analyses and task-specialist configurations.

Our empirical and diagnostic findings illustrate both the gains and the task-dependent nature of training-free looping in 3D vision (Figures~\ref{fig:radar_overview} and~\ref{fig:qualitative}).
VGGT raises HiRoom reconstruction F1 from 53.79 to 60.43 and MapAnything improves all reported reconstruction metrics across five benchmarks, alongside localized trade-offs under a single shared window and alternative task-specialist operating points.
Diagnostic analyses on HiRoom further link larger reconstruction gains in VGGT and Pi3 to higher intra-window patch-feature similarity, while VGGT trajectory comparisons connect effective windows to smaller downstream feature-direction deviations and show that consistent cache updates unlock windows spanning DPT readout layers.

Our contributions are threefold:
(1) an architecture-aware training-free looping framework for heterogeneous multi-view geometry backbones, extended to internal DPT readout layers via dual-stream cache updates;
(2) a five-model, nine-benchmark evaluation characterizing gains, task-dependent trade-offs, retrospective single-source transfer, and task-specialist alternatives;
and (3) diagnostic analyses on HiRoom connecting loop effectiveness to intra-window similarity, window placement, downstream feature-direction deviation, and consistent cache updates.

\section{Related Work}
\label{sec:related}
\begin{figure*}[t]
\centering
\includegraphics[width=\textwidth]{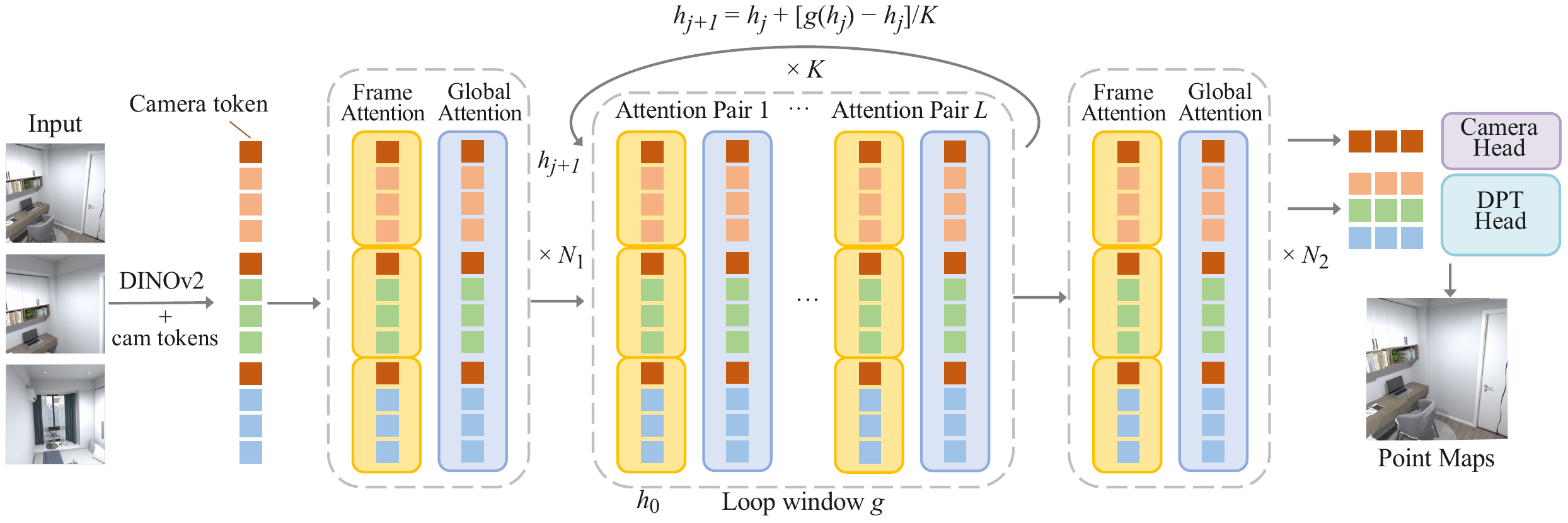}
\caption{Architecture-aware computation reuse in VGGT, with $N_1$ prefix pairs, $L$ window pairs, and $N_2$ suffix pairs. Starting from the prefix output $h_0$, the window $g=g_{s:e}$ is evaluated $K$ times with damped updates while preserving paired frame and global attention; $h_K$ enters the unchanged suffix and heads. When DPT feature-cache layers lie inside the window, Section~\ref{sec:method_cache} updates auxiliary frame and global cache states $(f_{c,j}, q_{c,j})$ in parallel (Equation~\ref{eq:cache}, omitted for clarity) so dense head readouts remain consistent with $h_K$.}
\label{fig:method}
\end{figure*}
\paragraph{Feed-forward 3D geometry models.}
Building on pairwise point-map prediction \citep{wang2024dust3r,leroy2024mast3r}, recent multi-view foundation models predict camera poses, dense depth, and 3D point maps in a single forward pass without per-scene optimization \citep{wang2025vggt,wang2025pi3,lin2025depthanything3,keetha2025mapanything,wang2026vggtomega}.
Despite sharing transformer backbones, these models differ substantially in how they structure multi-view interaction and readout: VGGT and Pi3 alternate within-view and global attention pairs; VGGT-$\Omega$ communicates across frames via special tokens; DA3 separates an initial single-view stage from a multi-view stage; and MapAnything fuses a scene-level scale token with per-view states.
Our work operates directly on these pretrained checkpoints with frozen weights.

\paragraph{Looped and recurrent transformers.}
Weight-tied recurrence across depth was pioneered by Universal Transformers \citep{dehghani2019universal} and extensively studied in looped transformers for programmable computation, iterative algorithm learning, length generalization, and latent reasoning \citep{giannou2023looped,yang2024looped,fan2025looped,saunshi2025reasoning,geiping2025scaling,zhu2025ouro}, as well as multi-view 3D reconstruction in D\'ej\`a View \citep{burzio2026dejaview}.
Unlike D\'ej\`a View, which trains a recurrent architecture, we introduce recurrence purely at inference time on existing feed-forward checkpoints.
Closer to our setting, \citet{chen2026trainingfree} and \citet{lys2026innerloop} apply training-free looping to frozen language models.
However, language models process a single token stream with a single final-layer readout, whereas multi-view geometry models couple frame and global attention, special or scale tokens, and multi-layer DPT caches \citep{ranftl2021dpt}.
Adapting damped looping thus requires preserving native graphs and cached dense features (Section~\ref{sec:method}).

\paragraph{Inference-time geometric interventions and layer probes.}
Easi3R \citep{chen2025easi3r} also intervenes in frozen geometry transformers at test time, targeting dynamic motion segmentation via attention reweighting rather than computation reuse.
Layer-wise geometric probes \citep{stary2025understanding} and block-recurrent vision-transformer analyses \citep{jacobs2025blockrecurrent} study how representations evolve across depth.
We link layer similarity and trajectory stability directly to training-free computation reuse.

\section{Architecture-Aware Computation Reuse}
\label{sec:method}
\subsection{Windowed damped updates}
Figure~\ref{fig:method} illustrates the computation flow for VGGT.
Let $g_{s:e}=B_e\circ\cdots\circ B_s$ be a contiguous, shape-preserving window of length $L=e-s+1$ in a pretrained backbone, with zero-based inclusive start and end indices $0\le s\le e$.
Depending on the architecture, each $B_i$ is either a paired within-view and cross-view block or a single physical transformer block.
The layers before and after the window form the prefix and suffix, followed by prediction heads that decode features into geometric outputs.
Given the prefix output $h_0$, standard inference applies the window once to produce $g_{s:e}(h_0)$.
Adapting damped computation reuse~\citep{chen2026trainingfree} to multi-view geometric backbones, we replace this unlooped window output with $K$ damped loop iterations:
\begin{equation}
h_{j+1}=h_j+\frac{g_{s:e}(h_j)-h_j}{K},
\qquad j=0,\ldots,K-1.
\label{eq:loop}
\end{equation}
Here $g_{s:e}(h_j)$ denotes the window proposal and $h_{j+1}$ the damped state, with the final state $h_K$ passed to the unchanged suffix and heads.
The wrapper introduces no learned parameters, adds $K-1$ extra window evaluations beyond the standard forward pass, and recovers native inference at $K=1$.
We denote this configuration by $[s,e]/K$ and vary $K\in\{2,3,4\}$ independently of $L$.

Equation~\ref{eq:loop} performs a forward-Euler step of size $1/K$ on the residual field $g_{s:e}(h)-h$ (equivalently, damped fixed-point updates).
We also consider an optional raw-endpoint anchor~\citep{chen2026trainingfree}, blending $h_K$ with the unlooped window output $a=g_{s:e}(h_0)$:
\begin{equation}
h_{\mathrm{out}}=\beta a+(1-\beta)h_K.
\label{eq:anchor}
\end{equation}
With blend weight $\beta\in[0,1]$, the suffix instead receives $h_{\mathrm{out}}$ (reducing to $h_K$ when $\beta=0$).
Our shared configurations use $\beta=0$, except for the VGGT-$\Omega$ setting in Table~\ref{tab:configs}.

\subsection{Respecting the native computation graph}
\label{sec:method_graph}
Unlike single-stream text transformers, geometric foundation models couple within-view and cross-view attention alongside specialized tokens, so each replayed unit $B_i$ must preserve the backbone's native computation graph.
For alternating-attention backbones, we treat complete attention pairs as indivisible units: frame and global pairs in VGGT and Pi3, frame and inter-frame pairs in VGGT-$\Omega$, and global and frame fusion pairs in MapAnything, where the scene-level scale token is updated jointly with per-view features.
By contrast, DA3 transitions from an initial single-view stage to a later multi-view stage; we respect this boundary by replaying single physical blocks within the single-view stage or complete pairs within the multi-view stage separately.

When an architecture relies on specialized control tokens, updating every token stream can cause unwanted drift.
For VGGT-$\Omega$, we therefore also consider token-restricted updates that hold unselected tokens fixed at their unlooped window outputs while keeping them active in native attention (Appendix~\ref{app:reproducibility}).

\subsection{Consistent intermediate caches for dense prediction}
\label{sec:method_cache}
Dense prediction heads such as DPT~\citep{ranftl2021dpt} read intermediate features cached from selected backbone layers (e.g., pairs $c\in\{4,11,17,23\}$ in VGGT), which we call \emph{feature-cache layers}.
Because Equation~\ref{eq:loop} applies damping only at the window output, naively saving an internal layer $c\in[s,e]$ during replay records an undamped proposal from a single pass rather than a damped state consistent with $h_K$ and the downstream suffix.
While this mismatch can be avoided by keeping feature-cache layers out of the window interior, we can also explicitly maintain consistent damped caches to unlock windows spanning internal feature-cache layers in VGGT (Section~\ref{sec:analysis}).

Specifically, let $(F_i,G_i)$ denote the frame and global outputs of pair $i$; because VGGT's DPT head reads their concatenation $[F_c,G_c]$ at each cached pair $c$, both streams must be updated consistently.
For a window starting at $s>0$ with prefix outputs $(F_{s-1},G_{s-1})$ (so $h_0=G_{s-1}$), we initialize auxiliary states $(f_{c,0},q_{c,0})=(F_{s-1},G_{s-1})$ for each cached pair $c\in[s,e]$ and update them in parallel with Equation~\ref{eq:loop}:
\begin{equation}
\begin{split}
f_{c,j+1}&=f_{c,j}+\bigl(F_c(h_j)-f_{c,j}\bigr)/K,\\
q_{c,j+1}&=q_{c,j}+\bigl(G_c(h_j)-q_{c,j}\bigr)/K,
\end{split}
\label{eq:cache}
\end{equation}
for $j=0,\ldots,K-1$, where $F_c(h_j)$ and $G_c(h_j)$ are the intermediate frame and global outputs at pair $c$ during the $j$-th window evaluation.
These auxiliary states never feed back into attention; after $K$ iterations, the DPT head reads $[f_{c,K},q_{c,K}]$ (with $q_{c,K}=h_K$ whenever $c=e$), ensuring that multi-scale dense readouts follow the same damped dynamics as the propagated backbone state.

\section{Multi-Model Geometric Evaluation}
\label{sec:experiments}
\subsection{Setup and reporting}
We evaluate official pretrained VGGT-1B, Pi3, DA3-Giant-1.1, MapAnything, and VGGT-$\Omega$-1B/512 checkpoints.
Inference is RGB-only; ground-truth (GT) geometry, poses, depths, and masks are used only for evaluation.
Raw and loop share checkpoints, native preprocessing and mixed precision, input order, sampling, and evaluators.
Checkpoint versions are detailed in Appendix~\ref{app:checkpoints}.

The suite combines benchmarks used in recent geometry evaluations:
HiRoom \citep{svlightverse2025,lin2025depthanything3},
ETH3D \citep{schops2017eth3d},
DTU \citep{aanaes2016dtu},
7Scenes \citep{shotton2013sevenscenes},
NRGBD \citep{azinovic2022neuralrgbd},
ScanNet++ \citep{yeshwanth2023scannetpp},
DyCheck \citep{gao2022dycheck},
Sintel \citep{butler2012sintel}, and
TUM-Dynamic \citep{sturm2012tumrgbd}.
We report camera AUC@3 and AUC@30, reconstruction F1 and distance, and depth $\delta_{1.25}$ and AbsRel wherever the corresponding task is defined.
Because native resolutions and reconstruction representations differ, we compare each model with its own raw inference.
Protocols follow DA3, Pi3, and VGGT-$\Omega$ evaluations \citep{lin2025depthanything3,wang2025pi3,wang2026vggtomega}, with model-specific details in Appendix~\ref{app:protocol}.

\paragraph{Evaluation configurations.}
In Section~\ref{sec:eval_main}, we evaluate one fixed shared configuration per model across all supported benchmarks (Tables~\ref{tab:aggregate}--\ref{tab:depth_dynamic}), keeping its window, $K$, token scope, and anchor constant with no per-dataset routing.
Section~\ref{sec:eval_transfer} tests cross-dataset generalization via retrospective single-source held-out transfer (Table~\ref{tab:single_source_transfer}; Appendix~\ref{app:search}).
Section~\ref{sec:eval_specialist} examines \emph{task-specialist configurations} tailored to camera pose, reconstruction, or depth with the same frozen weights, along with cross-pass output consistency (Appendices~\ref{app:specialists} and~\ref{app:multipass}).
Finally, Section~\ref{sec:eval_qual} presents qualitative reconstruction comparisons and iteration-wise geometric evolution.

\begin{table*}[t]
\centering
\caption{Shared configurations and cross-benchmark changes over raw inference (positive is better throughout). $\Delta$AUC, $\Delta$F1, and $\Delta\delta_{1.25}$ are in percentage points; Dist.\ red.\ and AbsRel red.\ denote mean per-benchmark relative error reduction (\%). All models update all tokens with $\beta=0$ except VGGT-$\Omega$ (patch-only, $\beta=0.125$); DA3 loops a single-view block, while the others loop attention pairs.}
\label{tab:configs}
\label{tab:aggregate}
\small
\setlength{\tabcolsep}{6pt}
\begin{tabular}{lcrrrrrr}
\toprule
 & & \multicolumn{2}{c}{Camera} & \multicolumn{2}{c}{Reconstruction} & \multicolumn{2}{c}{Depth} \\
Model & Window/$K$ & $\Delta$AUC@3 & $\Delta$AUC@30 & $\Delta$F1 & Dist.\ red.\ (\%) & $\Delta\delta_{1.25}$ & AbsRel red.\ (\%) \\
\midrule
VGGT & $[5,8]/4$ & +0.762 & +0.127 & +2.283 & +2.56 & +0.226 & +1.85 \\
Pi3 & $[4,8]/3$ & +0.206 & +0.182 & +0.334 & +2.01 & +0.088 & +1.35 \\
DA3 & $[6,6]/3$ & +0.319 & +0.113 & +0.263 & +0.54 & +0.059 & +0.06 \\
MapAnything & $[2,3]/2$ & +1.322 & +0.590 & +2.217 & +3.49 & $-0.060$ & +1.18 \\
VGGT-$\Omega$ & $[8,9]/4$ & +0.336 & +0.152 & -- & -- & +0.094 & +0.49 \\
\bottomrule
\end{tabular}
\end{table*}

\subsection{Main results across benchmarks and tasks}
\label{sec:eval_main}
\paragraph{Cross-benchmark aggregates.}
Table~\ref{tab:aggregate} summarizes the shared configurations and their cross-benchmark aggregate changes over raw inference.
VGGT, Pi3, and DA3 improve all six aggregate metrics, and VGGT-$\Omega$ improves all four of its applicable metrics.
MapAnything improves five of the six aggregate metrics, with a negligible $-0.060$ change in mean depth accuracy alongside a $+1.18\%$ AbsRel reduction.
Per-benchmark results appear in Tables~\ref{tab:pose_static}, \ref{tab:pose_video}, \ref{tab:reconstruction}, \ref{tab:depth_static}, and \ref{tab:depth_dynamic}.

\begin{table*}[t]
\centering
\caption{Static-scene camera pose. The same model-specific shared loop configuration is used in all main tables. VGGT-$\Omega$ uses 10-frame sampling on ETH3D and 7Scenes, unlike the other models' corresponding main-protocol cells. DTU$^{*}$ evaluates the released mixed-view pose subset (49 or 64 views per scene; Appendix~\ref{app:protocol}). In Tables~\ref{tab:pose_static}--\ref{tab:depth_dynamic}, small parentheses show $\Delta=\mathrm{loop}-\mathrm{raw}$ before score rounding: \textcolor[HTML]{16A34A}{green} indicates improvement, \textcolor[HTML]{DC2626}{red} regression, and black a delta rounding to zero.}
\label{tab:pose_static}
\setlength{\tabcolsep}{1pt}
\resizebox{\textwidth}{!}{%

}
\end{table*}

\begin{table*}[t]
\centering
\caption{Camera pose on NRGBD (static) and three dynamic benchmarks. All use the frozen 10-frame sampling protocol.}
\label{tab:pose_video}
\footnotesize
\setlength{\tabcolsep}{4pt}

%
\end{table*}

\begin{table*}[t]
\centering
\caption{Unposed reconstruction. F1 is a percentage; distance is Overall in cm for HiRoom, metres for ETH3D/7Scenes/ScanNet++, and Chamfer distance in mm for DTU (22 scans). ETH3D uses complete GT-depth support. A dash is not a zero.}
\label{tab:reconstruction}
\footnotesize
\setlength{\tabcolsep}{4pt}

%
\end{table*}

\begin{table}[t]
\centering
\caption{Static video-depth evaluation. Each sequence has a single L1 scale-and-shift fit; accuracy is shown in percent.}
\label{tab:depth_static}
\setlength{\tabcolsep}{3pt}
\resizebox{\linewidth}{!}{%
%
}
\end{table}

\begin{table}[t]
\centering
\caption{Dynamic video-depth evaluation, using the same frozen frames and evaluator for raw and loop.}
\label{tab:depth_dynamic}
\setlength{\tabcolsep}{3pt}
\resizebox{\linewidth}{!}{%
%
 \\
\addlinespace[0.5pt]
\bottomrule
\end{tabular}
}
\end{table}

\paragraph{Per-benchmark gains and variations.}
The largest dataset-level improvements appear in VGGT and MapAnything.
On HiRoom, VGGT improves camera AUC@3 from 48.19 to 52.20 and AUC@30 from 88.02 to 89.54 (Table~\ref{tab:pose_static}), while raising reconstruction F1 from 53.79 to 60.43 and reducing Overall distance from 10.33 to 8.88\,cm (Table~\ref{tab:reconstruction}).
In the reconstruction suite (Table~\ref{tab:reconstruction}), MapAnything improves all four F1 and all five distance entries: for example, ETH3D F1 increases from 57.76 to 61.13, 7Scenes F1 from 52.67 to 55.88, and DTU distance falls from 7.4345 to 6.8887\,mm.
Across the remaining backbones, Pi3 improves HiRoom F1 (80.41 to 81.70), DTU distance (2.9933 to 2.7703\,mm), 7Scenes, and ScanNet++ in Table~\ref{tab:reconstruction} while showing a localized drop on ETH3D F1 (73.41 to 71.36).
DA3's single-view block ($[6,6]/3$) raises HiRoom F1 (85.20 to 86.46; Table~\ref{tab:reconstruction}) with smaller shifts on dense depth (Tables~\ref{tab:depth_static} and~\ref{tab:depth_dynamic}), and VGGT-$\Omega$'s token-restricted setting achieves steady aggregate gains with minor cell-level variation across camera and depth benchmarks (Tables~\ref{tab:pose_static}, \ref{tab:pose_video}, \ref{tab:depth_static}, and~\ref{tab:depth_dynamic}).

\subsection{Cross-dataset transfer from a single source}
\label{sec:eval_transfer}
To examine whether a loop configuration calibrated on a single dataset generalizes across unseen benchmarks, Table~\ref{tab:single_source_transfer} reports a retrospective single-source held-out analysis: for each model, one configuration is selected solely from the listed source benchmark(s) and evaluated on all remaining benchmarks (Appendix~\ref{app:search}).
Selecting solely on HiRoom recovers VGGT's exact main-table configuration $[5,8]/4$ (22 improved vs.\ 12 degraded cells across the 8 held-out benchmarks; 20/8 when selecting on HiRoom+7Scenes).
Pi3 ($[4,12]/2$) and MapAnything ($[2,3]/4$) each achieve 25 improved vs.\ 10 degraded held-out cells from HiRoom alone, with MapAnything improving mean F1 by $+2.652$ and distance by $+3.49\%$.
For VGGT-$\Omega$ (which omits HiRoom in the main suite), selecting on NRGBD recovers its main-table $[8,9]/4$ setting (16/4 improved/degraded cells).
By contrast, single-source selection on HiRoom is less balanced for DA3 (improving held-out pose while trading off reconstruction and depth), showing that single-source transferability varies across architectures.

\begin{table*}[t]
\centering
\caption{Retrospective single-source held-out transfer. Each row selects one configuration solely from the listed source benchmark(s) and evaluates it on all remaining benchmarks. Cells report improved/degraded/tied metric counts and mean held-out changes: percentage points for AUC, F1, and $\delta_{1.25}$; relative error reduction (\%) for Dist.\ red.\ and AbsRel red.}
\label{tab:single_source_transfer}
\footnotesize
\setlength{\tabcolsep}{4pt}
\resizebox{\textwidth}{!}{%
\begin{tabular}{lllccccccc}
\toprule
Model & Source(s) & Selected config & Held-out (Imp./Deg./Tie) & $\Delta$AUC@3 & $\Delta$AUC@30 & $\Delta$F1 & Dist.\ red.\ (\%) & $\Delta\delta_{1.25}$ & AbsRel red.\ (\%) \\
\midrule
VGGT & HiRoom & $[5,8]/4$ & 8 (22/12/1) & +0.354 & $-0.047$ & +0.831 & $-0.29$ & +0.226 & +1.85 \\
VGGT & HiRoom+7Scenes & $[5,8]/4$ & 7 (20/8/1) & +0.409 & $-0.059$ & +1.265 & +0.20 & +0.261 & +2.22 \\
Pi3 & HiRoom & $[4,12]/2$ & 8 (25/10/0) & +0.614 & +0.497 & +1.242 & +1.46 & $-0.431$ & +3.26 \\
DA3 & HiRoom & $[2,8]/4$ (SV) & 8 (19/16/0) & +0.560 & +0.857 & $-1.108$ & $-3.01$ & $-0.077$ & $-2.06$ \\
MapAnything & HiRoom & $[2,3]/4$ & 8 (25/10/0) & +1.377 & +0.554 & +2.652 & +3.49 & $-0.106$ & +1.45 \\
VGGT-$\Omega$ & NRGBD & $[8,9]/4$ (patch, $\beta{=}0.125$) & 5 (16/4/0) & +0.272 & +0.165 & -- & -- & +0.113 & +0.50 \\
\bottomrule
\end{tabular}%
}
\end{table*}

\subsection{Task specialization and output consistency}
\label{sec:eval_specialist}
\begin{figure}[t]
\centering
\includegraphics[width=0.95\linewidth]{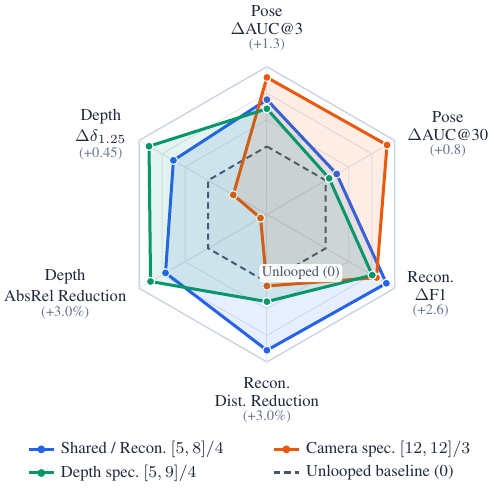}
\caption{VGGT cross-benchmark aggregate changes under its shared (reconstruction) configuration ($[5,8]/4$) versus camera ($[12,12]/3$) and depth ($[5,9]/4$) task-specialist windows (Appendix Table~\ref{tab:specialist_aggregate}). All axes are linear and anchored at the unlooped baseline (dashed inner hexagon, 0); outward indicates improvement.}
\label{fig:radar_specialist}
\end{figure}

\paragraph{Task-specialist operating points.}
Different geometric tasks respond best to different loop windows, so task-specialist configurations provide stronger single-task operating points using the same frozen weights (Appendix~\ref{app:specialists}).
As shown in Figure~\ref{fig:radar_specialist}, VGGT's camera specialist ($[12,12]/3$) raises mean $\Delta$AUC@3 and $\Delta$AUC@30 from $+0.762$ and $+0.127$ to $+1.125$ and $+0.710$ at the expense of depth accuracy, whereas its depth specialist ($[5,9]/4$) improves mean $\Delta\delta_{1.25}$ to $+0.385$ and AbsRel reduction to $+2.50\%$ while preserving gains across all other metrics (Appendix Table~\ref{tab:specialist_aggregate}).
For DA3, the camera specialist ($[2,10]/2$) raises $\Delta$AUC@3 and $\Delta$AUC@30 to $+1.381$ and $+0.672$, improving all 18 pose cells across static and dynamic benchmarks (Appendix Tables~\ref{tab:specialist_pose_static} and~\ref{tab:specialist_pose_video}), while its reconstruction specialist ($[0,1]/4$) increases mean distance reduction from $+0.54\%$ to $+2.34\%$.
VGGT-$\Omega$ exhibits analogous camera and depth specialist trade-offs; full specialist configurations and cross-task effects are reported in Appendix~\ref{app:specialists}.

\paragraph{Task routing and cross-pass consistency.}
A natural follow-up is whether outputs from separate specialists can be combined across passes.
Task-level routing, which takes each task's complete prediction from its dedicated specialist, improves VGGT's aggregate camera and depth metrics while preserving its reconstruction specialist's scores, at the cost of multiple forward passes (Appendix~\ref{app:multipass}).
In contrast, directly mixing geometric outputs across passes (combining cameras from the camera specialist with depths from the depth specialist for 3D reconstruction) worsens mean relative reconstruction distance by $0.23\%$ and $0.46\%$ relative to raw inference, compared with a $2.56\%$ reduction from the single unified reconstruction pass.
Stronger standalone camera poses and affine-aligned depths from different loop passes therefore do not automatically form consistent 3D geometry, underscoring the value of a single unified configuration when multi-task geometric consistency is required.

\subsection{Qualitative reconstruction examples}
\label{sec:eval_qual}
\begin{figure*}[t]
\centering
\includegraphics[width=0.88\textwidth]{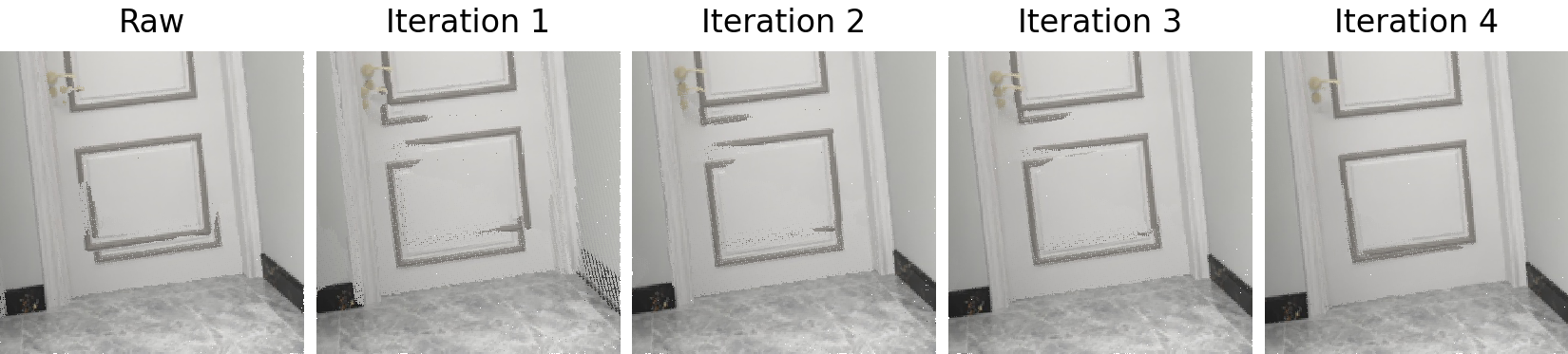}
\caption{VGGT reconstructions across loop iterations on HiRoom. Iterations 1--4 decode the damped states $h_1,\ldots,h_4$ of one $[5,8]/4$ run through the complete suffix and heads, alongside the unlooped baseline (Raw). After whole-scene similarity alignment, all panels share a viewing camera and rendering settings.}
\label{fig:iterations}
\end{figure*}

\paragraph{Cross-model visual comparisons.}
Figure~\ref{fig:qualitative} compares dense RGB point clouds from raw and loop inference across two HiRoom scenes, three DTU scans, and one ScanNet++ scene under identical observation cameras.
Across backbones, looping suppresses multi-view ghosting and sharpens fine surface structure: in the left group, VGGT removes the duplicated lower door moulding on HiRoom, Pi3 cleans up contours around the plush bird on DTU, and MapAnything substantially improves the structural coherence of the office cabinets on ScanNet++; in the right group, VGGT resolves double contours on the HiRoom notebook and desk items, DA3 sharpens the zipper and torso boundary on DTU, and VGGT-$\Omega$ improves window and cornice alignment on the DTU building facade.
VGGT and Pi3 use their shared configurations, whereas DA3 uses its reconstruction specialist and MapAnything and VGGT-$\Omega$ use task-aligned reconstruction configurations (input view counts appear in Appendix~\ref{app:reproducibility}).

\paragraph{Iteration-wise geometric refinement.}
Figure~\ref{fig:iterations} visualizes the reconstructions decoded from successive damped states $h_1,\ldots,h_4$ within a single VGGT $[5,8]/4$ run on HiRoom.
Because the damped update in Equation~\ref{eq:loop} advances from the window input $h_0$ with step size $1/K$, early intermediate states have only partially traversed the window and therefore exhibit transient displacement along the door-panel edges.
Across iterations, these displaced contours progressively contract and align, with the final state $h_4$ yielding a cleaner door moulding with substantially reduced double edges compared with raw inference.

\section{Where Does Repeated Computation Help?}
\label{sec:analysis}
The same loop update can improve one window and degrade another.
We investigate this behavior on HiRoom through three diagnostic questions: whether intra-window feature similarity and placement predict loop gains, how loop-induced changes evolve inside the loop versus in downstream layers, and whether windows can span internal DPT feature-cache layers.

\subsection{Intra-window similarity and placement}
A window may be easier to reuse if its layers already produce similar representations, making replay less disruptive to the pretrained feature manifold than repeating layers whose features shift sharply.
To test this hypothesis, we measure raw layer similarity as the mean cosine between corresponding patch tokens (averaged over patches and views within each scene, then equally across scenes).
For each window, we average this similarity over all distinct internal layer pairs and compare it against HiRoom F1 gain averaged over $K\in\{2,3,4\}$ (Appendix~\ref{app:similarity_controls}).

\begin{figure}[t]
\centering
\includegraphics[width=\linewidth]{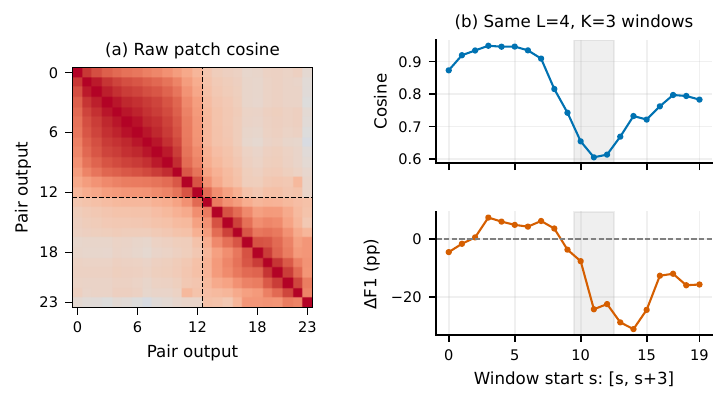}
\caption{VGGT similarity and F1 gain on HiRoom. (a) Raw patch-feature cosine between pair outputs; dashed lines mark the similarity-based division into earlier ($[0,12]$) and later ($[13,23]$) layers. (b) Within-window similarity (top) and F1 change (bottom) across all length-four windows $[s,s+3]$ at $K=3$ using consistent cache updates (Equation~\ref{eq:cache}). Gray shading marks windows spanning the similarity division.}
\label{fig:cache}
\end{figure}

\begin{figure}[t]
\centering
\includegraphics[width=0.94\linewidth]{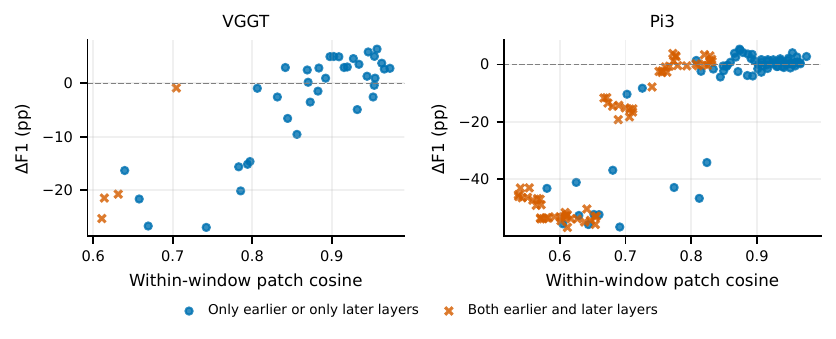}
\caption{Raw intra-window patch similarity versus HiRoom F1 gain. Each point is a window averaged over tested $K$. Blue circles lie wholly within earlier or later layers; orange crosses span both. Positive values improve over raw.}
\label{fig:scatter}
\end{figure}

Figure~\ref{fig:cache}(a) first reveals a two-block diagonal structure in VGGT's raw patch-feature similarity: earlier pairs $[0,12]$ resemble one another, as do later pairs $[13,23]$, with lower similarity across the boundary; Pi3 exhibits an analogous similarity-based division at $[0,10]$ and $[11,17]$ (Appendix~\ref{app:diagnostics}).
Figure~\ref{fig:scatter} relates this structure to loop gains.
In both VGGT and Pi3, higher within-window similarity tracks larger F1 gains: after controlling for window length and replay strategy, the stratified rank correlations are $0.717$ for VGGT and $0.675$ for Pi3 (Appendix~\ref{app:similarity_controls}).
Windows crossing the boundary (orange crosses) perform worse on average than matched non-crossing windows (blue circles), though a few Pi3 cross-boundary windows still improve over raw.

Scanning a fixed-length VGGT window across depth at $K=3$ (Figure~\ref{fig:cache}(b)) further shows that similarity and depth act jointly.
Early windows pair high similarity with positive gains, and both curves drop sharply across the boundary (Spearman $\rho=0.850$).
Deeper in the later layers, similarity partially rebounds while F1 remains below raw.
In Figure~\ref{fig:scatter}, 16 of 21 earlier-only VGGT windows improve F1 versus only 4 of 16 later-only windows (median $+2.95$ vs.\ $-12.10$).
Similarly, controlling for window start position in Pi3 lowers the rank correlation from $0.675$ to $0.320$, confirming that both placement and intra-window smoothness contribute.
In DA3's staged backbone, the controlled F1 correlation remains positive but weaker ($0.296$ across 91 windows, mostly in its single-view stage; Appendix~\ref{app:similarity_controls}).
Intra-window similarity is therefore most informative alongside placement in homogeneous alternating-attention backbones such as VGGT and Pi3.

\subsection{Loop dynamics and downstream feature drift}
\label{sec:trajectories}
\begin{figure}[t]
\centering
\includegraphics[width=\linewidth]{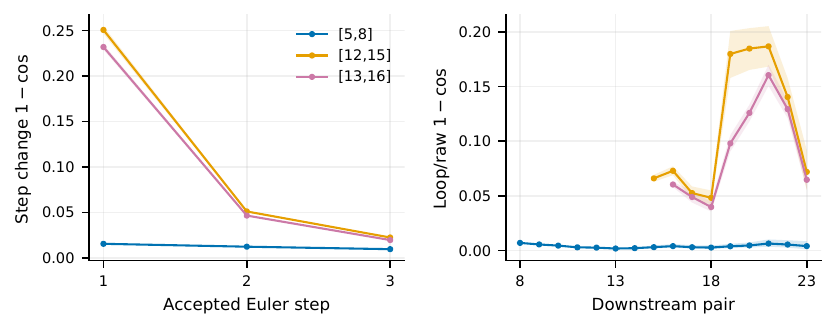}
\caption{VGGT patch-feature trajectories on HiRoom ($K=3$) for a beneficial window ($[5,8]$) versus two harmful windows ($[12,15]$, $[13,16]$). Left: per-iteration cosine distance $1-\cos(h_j, h_{j+1})$ between consecutive damped loop states ($j=0,1,2$, starting from the window input $h_0$). Right: downstream cosine distance $1-\cos(z_i^{\mathrm{loop}}, z_i^{\mathrm{raw}})$ relative to raw inference at each subsequent pair. Shading shows pointwise 95\% scene-bootstrap intervals.}
\label{fig:trajectories}
\end{figure}

To examine how beneficial and harmful loops differ in feature space, we compare three representative $L=4, K=3$ VGGT windows from Figure~\ref{fig:cache}(a) and~\ref{fig:cache}(b): the beneficial $[5,8]$ window and the harmful $[12,15]$ and $[13,16]$ windows.
In Figure~\ref{fig:trajectories}, we quantify patch-token directional deviation via cosine distance $1-\cos$ (which equals $0$ for aligned directions and increases as feature vectors rotate apart), tracking both the per-iteration change $1-\cos(h_j, h_{j+1})$ inside the loop and the downstream drift $1-\cos(z_i^{\mathrm{loop}}, z_i^{\mathrm{raw}})$ relative to raw inference at each subsequent pair $i$.

Because Equation~\ref{eq:loop} performs damped fixed-point updates, one might expect a well-behaved loop to be distinguished simply by whether its iterations converge (with step changes shrinking toward zero).
Figure~\ref{fig:trajectories} (left) shows that this is not enough: while $[5,8]$ makes small directional updates throughout, moving from $0.015$ to $0.010$, both harmful windows also settle rapidly after a large initial jump of $0.23$--$0.25$ at step~1 (e.g., $[13,16]$ drops from $0.232$ to $0.020$ despite losing $28.63$ F1 points relative to raw).
Settling to small late-step updates inside the loop therefore does not by itself indicate a beneficial update.

Instead, the decisive separation appears in the frozen downstream layers (Figure~\ref{fig:trajectories}, right).
While $[5,8]$ stays closely aligned with raw patch directions all the way to the final pair ($1-\cos = 0.004$), the directional shifts from $[12,15]$ and $[13,16]$ are amplified across later layers, peaking at $0.16$--$0.19$ around pairs 19--21 and remaining $0.072$ and $0.065$ at the final pair.
Effective training-free looping therefore acts as a gentle, direction-preserving refinement that remains compatible with the frozen suffix, rather than any iteration that merely converges internally (Appendix~\ref{app:diagnostics}).

\subsection{Looping across DPT feature-cache layers}
Section~\ref{sec:method_cache} introduced a dual-stream cache update (Equation~\ref{eq:cache}) so windows containing internal DPT feature-cache layers ($c\in\{4,11,17,23\}$ in VGGT) maintain damped caches consistent with the propagated state.
To test whether such windows remain effective in practice, we evaluate all 20 length-four VGGT windows $[s,s+3]$ ($s\in\{0,\ldots,19\}$, $K=3$) on HiRoom, spanning eight cache-free windows, six windows containing a cache layer strictly internally, and six starting or ending at one (Figure~\ref{fig:cache}(b)).
Window $[3,6]$, which contains cache layer 4 internally, achieves the highest F1 in the entire scan, reaching $61.11$ versus $53.79$ for raw ($+7.32$, paired scene-bootstrap 95\% interval $[1.59,13.63]$).
Position-matched local comparisons around $c\in\{4,11,17\}$ likewise show no uniform F1 penalty from internal cache layers ($+3.33$ F1 on average; Appendix~\ref{app:diagnostics}).
With consistent cache updates, DPT readout layers therefore do not need to be excluded from loop windows, though late windows still degrade F1 alongside the larger downstream feature-direction deviations observed above.

\section{Conclusion}
\label{sec:conclusion}
We present LoopVGGT, an architecture-aware framework for training-free looped inference across five heterogeneous geometric foundation models with frozen weights.
Fixed per-model configurations yield substantial geometric gains on several benchmarks alongside task-dependent trade-offs, while task-specialist configurations offer alternative task-specific operating points.
Diagnostic analyses on HiRoom associate VGGT and Pi3 reconstruction gains with intra-window feature similarity, link VGGT loop behavior to window position and downstream feature-direction deviation, and show that consistent dual-stream cache updates enable useful windows spanning DPT feature-cache layers.

\paragraph{Limitations.}
Because looped inference replays a contiguous window $K{-}1$ additional times, it increases computation relative to raw inference, and a fixed shared window can still exhibit task-dependent trade-offs across heterogeneous benchmarks. Furthermore, our held-out transfer study is retrospective on the completed sweep archive, and single-source selection is not uniformly robust across all five backbones (notably DA3). Developing lightweight, input-adaptive window and iteration selection remains open for future work.

{
    \small
    \bibliographystyle{ieeenat_fullname}
    \bibliography{main}
}

\clearpage
\maketitlesupplementary
\appendix
\raggedbottom

\section{Checkpoints and native geometry}
\label{app:checkpoints}
We evaluate official pretrained checkpoints for VGGT-1B, Pi3, DA3-Giant-1.1, MapAnything, and VGGT-$\Omega$-1B/512 (the originally released 512-resolution checkpoint \citep{vggtomega2026reproduction}).
Raw and loop comparisons share the same software environment, native precision, and preprocessing.
VGGT uses a BF16 backbone with FP32 geometry heads; DA3 uses its native 504 upper-bound preprocessing and first-reference camera-head path; and MapAnything uses native 518 resizing/cropping, image-only inference, and its scene-level scale token.
Where applicable, ray distances are converted to optical-axis depth during output adaptation.
Point-map reconstruction is used for Pi3 and MapAnything, whereas VGGT and DA3 use predicted cameras and depths with the corresponding fusion evaluator.

\section{Exact evaluation scope}
\label{app:protocol}
\begin{table*}[t]
\centering
\caption{Evaluated benchmark subsets. Scene counts denote the subsets evaluated in our experiments.}
\small
\begin{tabular}{lrll}
\toprule
Dataset & Scenes & Origin/content & Tasks \\
\midrule
HiRoom & 30 & Synthetic indoor, static & pose, reconstruction \\
ETH3D & 11 & Real indoor/outdoor, static & pose, reconstruction, depth \\
DTU$^{*}$ (pose) & 13 & Real objects, static & pose \\
DTU (reconstruction) & 22 & Real objects, static & reconstruction \\
7Scenes & 7 & Real indoor, static & pose, reconstruction, depth \\
NRGBD & 9 & Synthetic indoor, static & pose, depth \\
ScanNet++ & 20 & Real indoor, static & pose, reconstruction \\
DyCheck & 14 & Real, dynamic & pose, depth \\
Sintel & 14 & Synthetic, dynamic & pose, depth \\
TUM-Dynamic & 8 & Real indoor, dynamic & pose, depth \\
\bottomrule
\end{tabular}
\end{table*}

HiRoom uses all 400 released images across 30 scenes (10--23 per scene), ETH3D main pose and reconstruction use all 377 views, and ScanNet++ uses 100 fixed frames per scene (seed 42).
DTU reconstruction uses all 49 views across 22 scans (reference index 33) under the released DA3 evaluation settings, keeping depth-fusion and native point-map paths distinct.
The released DTU pose subset (DTU$^{*}$) contains 13 scenes with mixed view counts (seven 49-view and six 64-view scenes), evaluated identically for raw and loop.

\paragraph{Video sampling and camera evaluation.}
NRGBD and the three dynamic datasets use deterministic 10-frame random-order sampling (seed 42) following Pi3; for ETH3D depth, VGGT, Pi3, DA3, and MapAnything sample 10 frames per scene (seed 42) in original frame order, whereas VGGT-$\Omega$ uses random-order 10-frame sampling across all six of its benchmarks.
Camera pose accuracy is measured by averaging AUC@3 and AUC@30 of pairwise relative rotation and translation-direction angular errors across scenes.
On 7Scenes, VGGT, Pi3, DA3, and MapAnything use the DA3-BENCH evaluator (composing world-to-camera matrices as $E_i^{-1}E_j$), whereas VGGT-$\Omega$ uses the 10-frame VGGT evaluator ($E_iE_j^{-1}$); each raw and loop pair is always evaluated with the identical protocol.

\paragraph{Depth.}
A single affine transform $s\hat d+t$ is fitted by L1 loss jointly across all selected views of each sequence, and AbsRel and $\delta_{1.25}$ are weighted by valid-pixel count.
Valid depth ranges (in meters) are $(0,80)$ on ETH3D, $(0,10)$ on 7Scenes and NRGBD, $(0,3)$ on DyCheck, and $(0,70)$ on Sintel and TUM-Dynamic, preserving native crop/resize mappings and out-of-crop masks.

\paragraph{Reconstruction.}
Unposed reconstructions are aligned to ground truth via Umeyama--RANSAC trajectory alignment (seed 42).
VGGT and DA3 use scaled depths with TSDF fusion (and valid-depth backprojection on DTU), whereas Pi3 and MapAnything align native world-point maps directly.
The F1 threshold is 0.05\,m on HiRoom, 7Scenes, and ScanNet++, and 0.25\,m on ETH3D (evaluated across all 377 views over valid finite positive GT-depth masks).
DTU reports mean accuracy/completeness Chamfer distance in millimeters.
Because distance units differ across datasets, cross-benchmark aggregates average per-benchmark relative error reductions rather than raw distances.

\section{Per-benchmark result conventions}
\label{app:fullresults}
Each full-suite model covers 39 metric cells (18 camera, 9 reconstruction, 12 depth), while VGGT-$\Omega$ covers 24 camera and depth cells across six benchmarks.
Tables~\ref{tab:pose_static}--\ref{tab:depth_dynamic} report raw inference alongside the same fixed shared configuration from Table~\ref{tab:configs} without substituting task specialists.
Unsupported tasks are marked with dashes or omitted rows rather than zeros, and reported values are rounded only for display.

\section{Search coverage and selection details}
\label{app:search}
The native index axes are 24 attention pairs for VGGT and VGGT-$\Omega$, 18 attention pairs for Pi3, 40 physical blocks for DA3, and 16 fusion blocks for MapAnything.
Window endpoints are zero-based and inclusive.
For each model, we evaluate one fixed shared configuration across all supported benchmarks alongside task-specific configurations.

\begin{table*}[t]
\centering
\caption{Evaluated configuration spaces ($K\in\{2,3,4\}$ throughout).}
\small
\begin{tabular}{lp{10.5cm}r}
\toprule
Model & Admissible windows/design & Configurations \\
\midrule
VGGT & 61 pair windows excluding DPT feature-cache indices 4, 11, 17, 23 & 183 \\
Pi3 & All contiguous pair windows in 0--16; retain pair 17 & 459 \\
DA3 & 15 later pair windows and all 91 single-view windows in physical blocks 0--12 & 318 \\
MapAnything & 16 global and frame pair windows; feature-cache layers may end a window but cannot lie strictly inside it & 48 \\
VGGT-$\Omega$ & One-factor-at-a-time design across window, $K$, anchor, scope, and replay unit & 2,133 \\
\bottomrule
\end{tabular}
\end{table*}

Seven Pi3 configurations yield non-finite or empty outputs on at least one benchmark and are excluded from full-suite aggregates (452 complete), while retaining their valid HiRoom outputs in the HiRoom-only analysis.
DA3's 318 configurations consist of 45 multi-view and 273 initial single-view configurations.
MapAnything's 48 configurations include windows ending at a feature-cache layer, where the wrapper updates those features before head readout.
For VGGT-$\Omega$, starting from pair-based whole-window looping with all-token updates and $\beta=0$ across 183 window/$K$ pairs, the one-factor-at-a-time sweep varies one factor at a time (base, per-pair replay, five anchor strengths, and four nontrivial token scopes, giving $11\times 183$ settings, plus 60 frame-only and 60 inter-only controls for 2,133 settings in total), and combining it with joint anchor-and-scope configurations yields 2,333 distinct configurations.

\paragraph{Evaluated shared settings.}
The main-table shared configurations are VGGT $[5,8]/4$, Pi3 $[4,8]/3$, DA3 $[6,6]/3$ (single-view block), MapAnything $[2,3]/2$ (all tokens including the scene-level scale token), and VGGT-$\Omega$ patch-only $[8,9]/4$ with $\beta=0.125$.

\paragraph{Retrospective single-source held-out transfer.}
To examine whether a configuration calibrated on a single dataset generalizes across unseen benchmarks, main-paper Table~\ref{tab:single_source_transfer} reports a retrospective re-selection analysis where each configuration is chosen solely from one (or two) source benchmark(s) and evaluated on all remaining held-out benchmarks.
We note that this is a retrospective analysis on the completed sweep archive rather than untouched future test data.
For each source, the candidate pool includes all configurations valid on the source benchmark(s) plus \emph{raw} (184 for VGGT; 460/453 for Pi3 on HiRoom/HiRoom+7Scenes; 319 for DA3; 49 for MapAnything; and 2,334 for VGGT-$\Omega$).
Each unrounded source metric is ranked from 1 (best, averaging ties), and candidates are scored by a hierarchical mean rank weighted equally across metrics $\rightarrow$ tasks $\rightarrow$ source datasets:
\begin{equation}
S(c)=\frac{1}{|\mathcal D|}\sum_{d\in\mathcal D}\frac{1}{|\mathcal T_d|}\sum_{t\in\mathcal T_d}\frac{1}{|\mathcal M_{d,t}|}\sum_{m\in\mathcal M_{d,t}}\operatorname{rank}_{d,t,m}(c),
\end{equation}
selecting the lowest $S(c)$ (breaking ties by preferring \emph{raw}, then smaller $(K{-}1)(e{-}s{+}1)$, then configuration ID; excluding \emph{raw} leaves the primary winners unchanged).
We use HiRoom (and HiRoom+7Scenes, which adds depth) as sources for the four full-suite models, and NRGBD for VGGT-$\Omega$, which does not evaluate HiRoom in the main suite; held-out transfer outcomes are summarized in main-paper Table~\ref{tab:single_source_transfer}.

\section{Single-pass task-specialist results}
\label{app:specialists}
We report task-specialist configurations for VGGT, DA3, and VGGT-$\Omega$ alongside each model's shared setting.
Each row uses one complete model inference, with the same frozen weights as raw inference; all evaluated outputs come from that configuration.
Thus these tables expose both target-task gains and changes in the other tasks, before any multi-pass routing or mixing.
VGGT's shared setting is also its reconstruction specialist, so they occupy one row labeled shared/recon.

\paragraph{Configurations.}
Table~\ref{tab:specialist_aggregate} specifies every window and $K$, with zero-based inclusive indices.
VGGT windows use frame and global pairs and VGGT-$\Omega$ windows use frame and inter-frame pairs; all DA3 settings here use physical blocks in its single-view stage.
VGGT and DA3 update all tokens without an anchor.
For VGGT-$\Omega$, the shared setting updates patches with $\beta=0.125$, the camera specialist updates patches and registers with $\beta=0$, and the depth specialist updates patches and camera tokens with $\beta=0$.
These scopes restrict the loop update, not participation in the native attention operations.

\paragraph{Metric conventions.}
The six aggregate changes use equal weights across benchmarks: nine for each camera metric, four for F1, five for reconstruction distance, and six for each depth metric.
VGGT-$\Omega$ uses six benchmarks for each of its four applicable metrics.
Distance and AbsRel changes are computed as relative error reductions within each benchmark before averaging.
Tables~\ref{tab:specialist_pose_static}, \ref{tab:specialist_pose_video}, \ref{tab:specialist_reconstruction}, \ref{tab:specialist_depth_static}, and~\ref{tab:specialist_depth_dynamic} give the corresponding absolute scores and changes from raw.
The raw references and evaluation protocols are identical to the main tables.
Small parentheses show differences from raw computed before rounding; green denotes improvement, red regression, and black a difference rounding to zero.
Dashes denote unsupported tasks, not zero scores.

\begin{table*}[t]
\centering
\caption{Single-pass shared and task-specialist configurations: aggregate changes from each model's raw inference. AUC, F1 and depth accuracy changes are percentage points; Dist./AbsRel are the mean per-benchmark relative error reductions (percent). Positive values indicate improvement in every column. All supported tasks are shown for every configuration, not only its selection task. VGGT's shared and reconstruction-specialist configuration is identical.}
\label{tab:specialist_aggregate}
\footnotesize
\setlength{\tabcolsep}{5pt}
\begin{tabular}{ccccccccc}
\toprule
Model & Method & Window/$K$ & $\Delta$AUC3 & $\Delta$AUC30 & $\Delta$F1 & Dist. red. & $\Delta\delta$ & AbsRel red.\\
\midrule
VGGT & shared/recon. & $[5,8]/4$ & \textcolor[HTML]{16A34A}{+0.762} & \textcolor[HTML]{16A34A}{+0.127} & \textcolor[HTML]{16A34A}{+2.283} & \textcolor[HTML]{16A34A}{+2.56} & \textcolor[HTML]{16A34A}{+0.226} & \textcolor[HTML]{16A34A}{+1.85}\\
VGGT & camera & $[12,12]/3$ & \textcolor[HTML]{16A34A}{+1.125} & \textcolor[HTML]{16A34A}{+0.710} & \textcolor[HTML]{16A34A}{+1.913} & \textcolor[HTML]{16A34A}{+0.14} & \textcolor[HTML]{DC2626}{-0.164} & \textcolor[HTML]{DC2626}{-2.68}\\
VGGT & depth & $[5,9]/4$ & \textcolor[HTML]{16A34A}{+0.614} & \textcolor[HTML]{16A34A}{+0.040} & \textcolor[HTML]{16A34A}{+1.752} & \textcolor[HTML]{16A34A}{+0.73} & \textcolor[HTML]{16A34A}{+0.385} & \textcolor[HTML]{16A34A}{+2.50}\\
\midrule
DA3 & shared & $[6,6]/3$ & \textcolor[HTML]{16A34A}{+0.319} & \textcolor[HTML]{16A34A}{+0.113} & \textcolor[HTML]{16A34A}{+0.263} & \textcolor[HTML]{16A34A}{+0.54} & \textcolor[HTML]{16A34A}{+0.059} & \textcolor[HTML]{16A34A}{+0.06}\\
DA3 & camera & $[2,10]/2$ & \textcolor[HTML]{16A34A}{+1.381} & \textcolor[HTML]{16A34A}{+0.672} & \textcolor[HTML]{16A34A}{+0.105} & \textcolor[HTML]{16A34A}{+0.20} & \textcolor[HTML]{DC2626}{-0.385} & \textcolor[HTML]{DC2626}{-0.91}\\
DA3 & recon. & $[0,1]/4$ & \textcolor[HTML]{DC2626}{-0.016} & \textcolor[HTML]{DC2626}{-0.239} & \textcolor[HTML]{16A34A}{+0.404} & \textcolor[HTML]{16A34A}{+2.34} & \textcolor[HTML]{DC2626}{-0.683} & \textcolor[HTML]{DC2626}{-0.52}\\
DA3 & depth & $[2,12]/2$ & \textcolor[HTML]{DC2626}{-0.041} & \textcolor[HTML]{16A34A}{+0.135} & \textcolor[HTML]{DC2626}{-0.947} & \textcolor[HTML]{DC2626}{-3.72} & \textcolor[HTML]{16A34A}{+0.089} & \textcolor[HTML]{16A34A}{+0.70}\\
\bottomrule
\end{tabular}
\end{table*}

\begin{table*}[t]
\centering
\caption{Single-pass specialists: static-scene camera pose. Configurations are defined in Table~\ref{tab:specialist_aggregate}; raw references and protocols match Table~\ref{tab:pose_static}. Each row uses complete outputs from one model inference. DTU$^{*}$ retains the mixed-view pose subset; VGGT-$\Omega$ uses 10-frame sampling.}
\label{tab:specialist_pose_static}
\setlength{\tabcolsep}{1pt}
\resizebox{\textwidth}{!}{%
%
}
\end{table*}

\begin{table*}[t]
\centering
\caption{Single-pass specialists: camera pose on NRGBD and dynamic scenes. Configurations are defined in Table~\ref{tab:specialist_aggregate}; raw references and protocols match Table~\ref{tab:pose_video}. Each row uses complete outputs from one model inference.}
\label{tab:specialist_pose_video}
\footnotesize
\setlength{\tabcolsep}{4pt}
%
\end{table*}

\begin{table*}[t]
\centering
\caption{Single-pass specialists: unposed reconstruction. Configurations are defined in Table~\ref{tab:specialist_aggregate}; raw references and protocols match Table~\ref{tab:reconstruction}. Each row uses complete outputs from one model inference. F1 is in percent; distances are cm on HiRoom, mm on DTU, and metres elsewhere. ETH3D uses complete GT-depth support.}
\label{tab:specialist_reconstruction}
\footnotesize
\setlength{\tabcolsep}{4pt}
%
\end{table*}

\begin{table*}[t]
\centering
\caption{Single-pass specialists: static-scene depth. Configurations are defined in Table~\ref{tab:specialist_aggregate}; raw references and protocols match Table~\ref{tab:depth_static}. Each row uses complete outputs from one model inference. $\delta_{1.25}$ is in percent; AbsRel is unscaled. Depth uses one scale-and-shift alignment per sequence.}
\label{tab:specialist_depth_static}
\setlength{\tabcolsep}{3pt}
\small
%
\end{table*}

\begin{table*}[t]
\centering
\caption{Single-pass specialists: dynamic-scene depth. Configurations are defined in Table~\ref{tab:specialist_aggregate}; raw references and protocols match Table~\ref{tab:depth_dynamic}. Each row uses complete outputs from one model inference. $\delta_{1.25}$ is in percent; AbsRel is unscaled. Depth uses one scale-and-shift alignment per sequence.}
\label{tab:specialist_depth_dynamic}
\setlength{\tabcolsep}{3pt}
\small
%
\\
\addlinespace[3pt]
\bottomrule
\end{tabular}
\end{table*}

\section{Task routing versus mixing geometric outputs}
\label{app:multipass}
Define VGGT specialists $\mathcal P=[12,12]/3$ for camera, $\mathcal R=[5,8]/4$ for reconstruction, and $\mathcal D=[5,9]/4$ for depth.
$E,I,Z$ denote extrinsics, intrinsics, and depth.
Task routing uses each pass's complete relevant output; geometric mixing assembles a reconstruction from different passes.
All rows retain all five reconstruction benchmarks.
The mixed-output row labels specify only the reconstruction inputs: every such row reports camera metrics from $\mathcal P$ and depth metrics from $\mathcal D$.
Consequently, $E_{\mathcal P}+(I,Z)_{\mathcal R}$ requires a third pass $\mathcal D$ for its independently evaluated depth output, even though its reconstruction uses only $\mathcal P$ and $\mathcal R$.

\begin{table*}[t]
\centering
\caption{Multi-pass VGGT aggregate metrics. Positive entries denote improvement from raw. Dist./AbsRel are relative reductions (\%); other changes are percentage points. Passes count full model executions, not internal $K$.}
\setlength{\tabcolsep}{3pt}

\small
\begin{tabular}{lrrrrrrr}
\toprule
Method & Passes & $\Delta$AUC3 & $\Delta$AUC30 & $\Delta$F1 & Dist.\ red. & $\Delta\delta$ & AbsRel red.\\
\midrule
Unified $\mathcal R$ & 1 & +.762 & +.127 & +2.283 & +2.56 & +.226 & +1.85\\
$E_{\mathcal P}+(I,Z)_{\mathcal D}$ & 2 & +1.125 & +.710 & +.756 & -.23 & +.385 & +2.50\\
$(E,I)_{\mathcal P}+Z_{\mathcal D}$ & 2 & +1.125 & +.710 & +.642 & -.46 & +.385 & +2.50\\
$E_{\mathcal P}+(I,Z)_{\mathcal R}$ & 3 & +1.125 & +.710 & -.094 & -1.08 & +.385 & +2.50\\
Task-routed $\mathcal P/\mathcal R/\mathcal D$ & 3 & +1.125 & +.710 & +2.283 & +2.56 & +.385 & +2.50\\
\bottomrule
\end{tabular}
\end{table*}

The three-pass task-routed system improves 28 of 39 cells from raw, compared with 26 for unified $\mathcal R$.
Against unified $\mathcal R$ directly, routing has 20 wins, ten losses, and nine ties.
Thus task specialization improves aggregate trade-offs without eliminating all regressions or preserving the cost of a single pass.
The mixed-output results are consistent with a cross-pass calibration/geometry mismatch, but do not independently identify its causal source.

\section{Additional representation evidence}
\label{app:diagnostics}
\paragraph{Similarity conventions and controls.}
\label{app:similarity_controls}
Raw features are collected on all 30 HiRoom scenes (400 images).
Let $z_{i,v,t}$ denote token $t$ in view $v$ after backbone unit $i$, and let $T$ select a token group.
The scene-level similarity is
\begin{equation}
S_T(i,j)=\mathbb{E}_{v,t\in T}
\frac{\langle z_{i,v,t},z_{j,v,t}\rangle}
{\|z_{i,v,t}\|_2\|z_{j,v,t}\|_2}.
\label{eq:similarity}
\end{equation}
We average scenes equally, then average distinct layer pairs within a window; single-layer windows have no defined internal similarity.
The correlation analysis covers 960 configurations across VGGT (183), Pi3 (459), and DA3 (318); MapAnything (which has only 15 multi-layer candidate windows) and VGGT-$\Omega$ (not evaluated on HiRoom) are omitted from this stratified correlation.
All raw matrices use corresponding-token cosine within a model; token grids across different native model resolutions are not matched to one another.
Features are normalized in FP32 and scene reductions use high-precision accumulation.
A pooled-patch cosine is a different statistic from mean corresponding-patch cosine.
We measure each backbone's native token groups: patches throughout; camera and registers in VGGT; registers in Pi3; and the CLS and camera slots in DA3's respective stages.

For each window we first average its gains over the complete set $K=2,3,4$.
We then group windows by their length and update strategy, with DA3's single-view and multi-view windows forming separate strategy groups.
Windows without defined internal similarity and groups containing fewer than two windows are excluded.
Within a group of size $n$, similarity and gain are ranked separately, with average ranks for ties, and transformed to $\tilde r=(r-(n+1)/2)/n$.
The reported correlation is the Pearson correlation of these transformed ranks concatenated across groups.
For the window-start control, we transform start-index ranks in the same way to form a vector $p$, subtract the fitted component $p(p^\top x)/(p^\top p)$ from each rank vector $x$, and correlate the two residual vectors.
This removes one linear start-rank effect, not every dependence on depth position.
The additional window-start adjustment reduces the VGGT and Pi3 F1 correlations from 0.717/0.675 to 0.664/0.320.

\begin{table*}[t]
\centering
\caption{Length- and strategy-controlled rank correlation between raw patch similarity and HiRoom gain, averaging $K=2,3,4$ within each window. Overall is relative distance reduction. These are descriptive correlations of overlapping windows rather than independent-sample significance tests.}
\small
\begin{tabular}{lrrrrr}
\toprule
Model & AUC3 & AUC30 & F1 & Overall & Used windows\\
\midrule
VGGT & .876 & .761 & .717 & .780 & 40\\
Pi3 & .568 & .600 & .675 & .646 & 135\\
DA3 & .172 & .322 & .296 & .190 & 91\\
\bottomrule
\end{tabular}
\end{table*}

The division into earlier and later layers shown in Figures~\ref{fig:scatter} and~\ref{fig:cache} uses the recursive maximum-cut procedure released by \citet{jacobs2025blockrecurrent}, with $\alpha=0.10$.
It yields VGGT $[0,12]/[13,23]$ and Pi3 $[0,10]/[11,17]$ for patch features; these are similarity-based divisions, not architectural feature-cache locations.
It forms edge weights $D_{ij}=1-S(i,j)$ and recursively splits a current group by its maximum-weight cut only when that cut exceeds the fixed threshold $\epsilon=\alpha\sum_{i<j}D_{ij}$, computed once from the full matrix.
Groups with fewer than four layers are not split further; child groups are not constrained to be contiguous.
Increasing $\alpha$ stops finer splits without changing the cosine matrix.
At a finer Pi3 partition, even the useful main-table window spans two resulting groups.
The classification therefore depends on the partitioning threshold.
Under the displayed division, Pi3 window $[8,11]/3$ contains earlier and later layers and improves F1, while the neighboring $[7,10]/3$ lies entirely in the earlier stage and does not.
The counts of improving earlier/later VGGT windows and their median gains in the main text use the mean gain over $K=2,3,4$ for each window.

\paragraph{Raw direction and norms.}
Here G$\ell$ denotes the global-attention output of VGGT pair $\ell$.
Mean patch cosine is 0.96744 between adjacent outputs inside G5--G8, but 0.59961 from G22 to G23.
Thus the small raw directional changes in G5--G8 are local to that part of the backbone.
In VGGT, adjacent G5--G8 camera/register cosines are .99647/.99636.
The same token groups have G12-versus-G23 cosines .10599/.09225 and G22-versus-G23 cosines .64124/.63389.
Mean per-scene norm ratios G23/G0 are 66.14, 66.70, and 16.60 for camera, registers, and patches.
A norm ratio, an adjacent direction change, and alignment to a distant endpoint are different observables.

\paragraph{Loop traces and uncertainty.}
Five VGGT windows $[5,8]$, $[8,11]$, $[9,12]$, $[12,15]$, and $[13,16]$, all $K=3$ with all-token updates, have complete 30-scene trajectories.
Observers distinguish replay proposals from damped states and verify unchanged native predictions.
For successive updated states we measure $1-\cos(h_j,h_{j+1})$; at each later pair $i$ we measure $1-\cos(z^{\mathrm{loop}}_{i,v,t},z^{\mathrm{raw}}_{i,v,t})$.
The plotted measurements use the patch-token components of these states, averaging corresponding tokens within each scene and then weighting scenes equally.
For patch tokens, the consecutive-state difference decreases from 0.0154 to 0.0095 in $[5,8]$, and from 0.2318 to 0.0195 in $[13,16]$, whose F1 is 28.63 points below raw.
The final-pair loop/raw differences are 0.00405, 0.07198, and 0.06475 for $[5,8]$, $[12,15]$, and $[13,16]$, respectively.
Later-layer feature norms can approach raw norms while their directions remain different.
The plotted intervals use 2,000 paired scene-bootstrap draws, seed 42, and are pointwise 95\% percentile intervals rather than simultaneous bands.
The cache scan intervals are exploratory and are not adjusted for selecting the best window or for multiple comparisons.

\paragraph{Windows containing DPT feature-cache layers.}
The scan evaluates all 20 length-four VGGT windows starting at pairs 0 through 19 on HiRoom, with $K=3$ and the paired feature-cache update in Equation~\ref{eq:cache}.
The DPT feature-cache indices are 4, 11, 17, and 23; no scanned window includes the final index 23.
Eight windows contain no feature-cache layer, three end at one, six contain one strictly internally, and three start at one.
The Spearman correlations between window similarity and AUC@3, AUC@30, F1, and distance improvement are 0.860, 0.895, 0.850, and 0.821.
These overlapping windows provide descriptive associations rather than independent samples.
The best-F1 window $[3,6]$ does not outperform $[5,8]/3$ on every task: the latter has better pose and Overall.
The strict-internal versus no-cache F1 difference is $+2.24$ points across the full groups, but the Overall difference is $+1.12$\,cm (worse).
The local control compares starts $c-2,c-1$ (a feature-cache layer strictly inside the window) with $c-4,c+1$ (no feature-cache layer) around each feature-cache index $c\in\{4,11,17\}$.
We average the two windows on each side within each scene and weight the three feature-cache locations equally, giving both sides the same mean start $c-1.5$ at each location.
This matches mean position but not all nonlinear effects of position or of the similarity-based division into earlier and later layers.
This local matching control yields $+3.33$ F1 points.
Separately, excluding windows that contain both earlier and later layers from the full-group comparison yields $+0.51$ with an interval spanning zero; AUC@30 is lower by 1.37 points and Overall worse by 1.67\,cm.
The differences depend on metric and control; absence of a uniform penalty is not evidence of equivalence.
VGGT's camera head reads the final backbone features, while the auxiliary feature-cache states supply the dense heads and never feed back into attention.
Changing only the auxiliary readout policy for a fixed window that excludes the final feature-cache layer therefore leaves camera predictions unchanged.

\section{Additional implementation details}
\label{app:reproducibility}
\paragraph{Token-restricted updates.}
In VGGT-$\Omega$, a binary token mask $M$ restricts which tokens accept the loop update.
Starting from the native window input $h_0$, each iteration forms a candidate $u_j$ and merges it with the unlooped window output $a=g_{s:e}(h_0)$:
\begin{equation}
\begin{split}
u_j&=h_j+\bigl(g_{s:e}(h_j)-h_j\bigr)/K,\\
h_{j+1}&=M\odot u_j+(1-M)\odot a.
\end{split}
\label{eq:token_scope}
\end{equation}
After $K$ masked updates, we apply the anchor blend in Equation~\ref{eq:anchor}, then merge its output with $a$ using the same mask.
Unselected tokens retain their unlooped window outputs while participating in every attention operation where the native architecture includes them.
Thus patch-only updates do not remove camera or register tokens from attention.

\paragraph{Replay granularity.}
In the VGGT-$\Omega$ search space (Appendix~\ref{app:search}), whole-window mode loops $g_{s:e}$ jointly, per-pair mode applies $K$ updates sequentially at each pair, and frame-only/inter-only controls loop one sublayer of a single pair in native order.

\paragraph{Rendering inputs.}
The qualitative reconstructions in Figure~\ref{fig:qualitative} use all selected inputs: 10 views per HiRoom scene, 49 per DTU scan, and 100 for the ScanNet++ scene.

\end{document}